\documentclass{article}
\usepackage{spconf,amsmath,graphicx}

\usepackage{multirow}
\usepackage{makecell}
\usepackage{CJKutf8}

\usepackage{booktabs}
\usepackage{url}

\title{Evaluating ChatGPT on Medical Information Extraction Tasks: Performance, Explainability and Beyond}
\twoauthors
{Liz Li}
{DataSelect AI\\
Xuhui, Shanghai, China}
{Wei Zhu\sthanks{Email: michaelwzhu91@gmail.com}}
{University of Hong Kong\\
Hong Kong, HK, China}

\begin{document}
%\ninept
%
\maketitle
\begin{abstract}
Large Language Models (LLMs) like ChatGPT have demonstrated amazing capabilities in comprehending user intents and generate reasonable and useful responses. Beside their ability to chat, their capabilities in various natural language processing (NLP) tasks are of interest to the research community. In this paper, we focus on assessing the overall ability of ChatGPT in 4 different medical information extraction (MedIE) tasks across 6 benchmark datasets. We present the systematically analysis by measuring ChatGPT’s performance, explainability, confidence, faithfulness, and uncertainty. Our experiments reveal that: (a) ChatGPT's performance scores on MedIE tasks fall behind those of the fine-tuned baseline models. (b) ChatGPT can provide high-quality explanations for its decisions, however, ChatGPT is over-confident in its predcitions. (c) ChatGPT demonstrates a high level of faithfulness to the original text in the majority of cases. (d) The uncertainty in generation causes uncertainty in information extraction results, thus may hinder its applications in MedIE tasks.

\end{abstract}

\section{Introduction}

Large Language Models (LLMs) (e.g., GPT-3 \cite{brown2020language}, LaMDA \cite{thoppilan2022lamda} and PaLM \cite{chowdhery2022palm}, GPT-4 \cite{2023arXiv230308774O}) are the main forces in revolutionizing and advancing the field of Natural Language Processing (NLP) \cite{achiam2023gpt4,anil2023palm2,singhal2023large,singhal2023towards,nori2023capabilities,huang2023c,li2023cmmlu,Cui2023UltraFeedbackBL,wang2024ts,yue2023-TCMEB,Zhang2023LearnedAA,2023arXiv230318223Z,Xu2023ParameterEfficientFM,Ding2022DeltaTA,Xin2024ParameterEfficientFF,qin2023chatgpt,PromptCBLUE,text2dt_shared_task,Text2dt,zhu_etal_2021_paht,Li2023UnifiedDR,Zhu2023BADGESU,Zhang2023LECOIE,Zhu2023OverviewOT,guo-etal-2021-global,zhu-etal-2021-discovering,Zheng2023CandidateSF,info:doi/10.2196/17653,Zhang2023NAGNERAU,Zhang2023FastNERSU,Wang2023MultitaskEL,Zhu2019TheDS,zhu2021leebert,Zhang2021AutomaticSN,Wang2020MiningIH,li2025ft,leong2025amas,zhang2025time,yin2024machine,zhu2026mrag}. LLMs are known for its amazing abilities: (a) instruction and demonstration learning. With a proper instruction of the task at hand and a few demonstrations \cite{ouyang2022training, chung2022scaling, wang2022super, min2022rethinking, li2023unified}, the LLMs can perform tasks they are yet trained for. (b) chain-of-thought (CoT) \cite{wei2022chain,wang2022self,kojima2022large} capabilities. LLMs have demonstrated great performances in solving a wide range of reasoning tasks by reasoning step-by-step. This reasoning capability is not observed in smaller models, thus is referred as an emergent ability in the literature \cite{wei2022emergent,tian2024opportunities,hersh2024search,zhu2024iapt,zhu-tan-2023-spt,Liu2022FewShotPF,xie2024pedro,Cui2023UltraFeedbackBL,zheng2024nat4at,zhu2023acf,gao2023f,zuo-etal-2022-continually,zhang-etal-2022-pcee,sun-etal-2022-simple,zhu-etal-2021-gaml,Zhu2021MVPBERTMP,li-etal-2019-pingan,zhu2019panlp,zhu2019dr,zhou2019analysis,zhang2025time,wang2025ts,liu2025parameter,yi2024drum,tian2024fanlora}.

ChatGPT\footnote{https://chat.openai.com/} has became the most popular AI based product within a short time span after its launch \cite{2023arXiv230213795L}. As a LLM, ChatGPT is known for its impressive ability to comprehend the user intents in the queries and repond with human-like contents. ChatGPT is built on the GPT-3 model series \cite{brown2020language, 2020arXiv200901325S, ouyang2022training} using reinforcement learning from human feed-back (RLHF) \cite{christiano2017deep, li2019human} and high-quality human-annotated datasets of user queries and responses. Apart from its ability to chat with humans, ChatGPT has many other aspects that are of interest to NLP researchers. A series of research work have emerged, investigating the capabilities of ChatGPT for various NLP tasks, like sentiment analysis \cite{susnjak2023applying,wang2023chatgpt}, natural language inference \cite{2023arXiv230206476Q}, machine reading comprehension \cite{yang2023harnessing}. The capabilities of ChatGPT as a general NLP task solver have been preliminarily explored in the above research and valuable insights are drawn. There are also work addressing the potential impact of ChatGPT on the society and human every-day life \cite{haque2022think, zhuo2023exploring, susnjak2022chatgpt,basic2023better}.

\begin{figure*}
	\centering
	\includegraphics[width=0.95\textwidth]{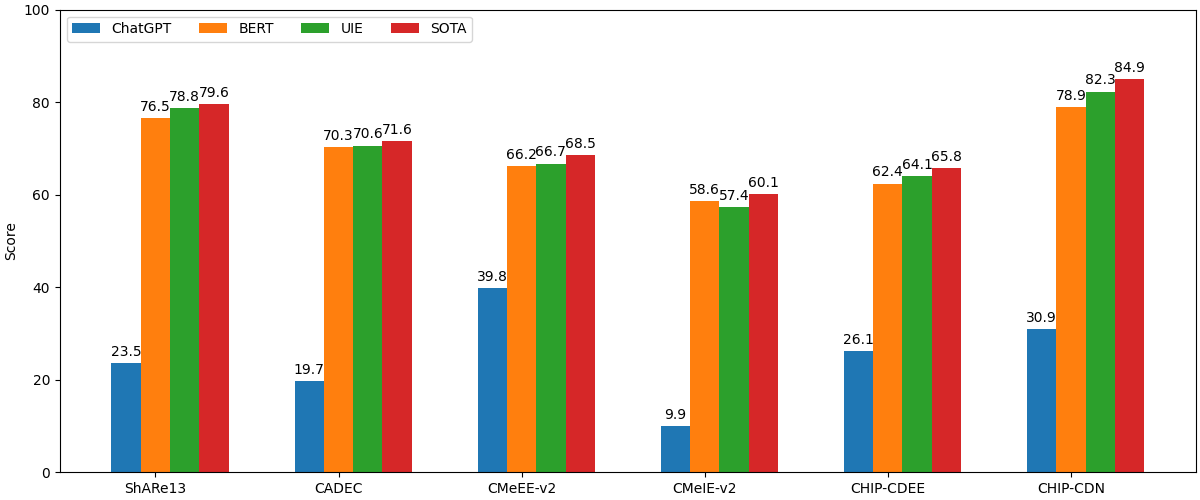}
	\caption{Performance comparisons of ChatGPT and the baseline models.}
	\label{fig:teaser}
\end{figure*}

In this work, we conduct a comprehensive analysis of ChatGPT's abilities in medical information extraction (MedIE) tasks. MedIE requires the model to have a deep understanding of the medical texts, and extract structured factual knowledge of diversified forms \cite{hahn2020medical, wang2018clinical, paolini2021structured, zhu-etal-2021-discovering, zhu2021autotrans}, thus it is an ideal test-bed for evaluating ChatGPT's capabilities. We conduct our experiments and analysis based on 6 datasets belonging to 4 fine-grained MedIE tasks. Through our experiments, we find that with carefully designed task instructions and demonstrations, ChatGPT still has clear performance gaps compared with fine-tuned baseline models, as depicted in Figure \ref{fig:teaser}. In addition to the performance evaluation, we also assess the explainability, faithfulness, confidence and uncertainty of the ChatGPT's responses through both automatic evaluations and manual evaluations by domain experts. 

Our contributions are summarized as follows:
\begin{itemize}
\item We comprehensively evaluate the overall performance of ChatGPT on various MedIE tasks and compare it with other popular models. 
\item To assess the overall ability of ChatGPT, we conduct a systematic evaluation from five dimensions: performance, explainability, faithfulness, confidence and uncertainty. Our codes are publicly available for future research.

\end{itemize}

\section{Related work}

\subsection{Pretrained Large Language Models}

% (e.g., GPT-3 \cite{brown2020language}, LaMDA \cite{thoppilan2022lamda} and PaLM \cite{chowdhery2022palm}, GPT-4 \cite{2023arXiv230308774O})

Recently, both the research field and the industry has seen the raise of large language models (LLMs). LLMs typically contain more than ten billion or even a hundred billion parameters, such as GPT-4 \cite{2023arXiv230308774O}, GPT-3 \cite{brown2020language}, Gopher \cite{rae2021scaling}, Chinchilla \cite{2022arXiv220315556H}, PaLM \cite{chowdhery2022palm}, Megatron-turing-NLG \cite{2022arXiv220111990S}, etc. Scaling up the size of LLMs has resulted in a series of interesting observations: (a) scaling up the model size brings impressive abilities in few-shot and zero-shot learning scenarios, such as producing reasonable results with very few task demonstrations or task descriptions \cite{brown2020language, chowdhery2022palm}. (b) scaling up the number of tasks in fine-tuning LLMs with instruction tuning improves the generalizability on unseen tasks \cite{longpre2023flan}. (c) Emergent capabilities are unlocked when the model size is scaled up to hundreds of billions, which were not observed in smaller models \cite{wei2022emergent}. Notably, the chain of thought capability is observed in LLMs \cite{wei2022chain, wang2022self, kojima2022large, mahowald2023dissociating}.

With its impressive ability to understand user in-tent and generate human-like responses, ChatGPT has become the most popular language model within a short period of time after its lanuch. It is trained on the GPT family \cite{brown2020language, 2020arXiv200901325S, ouyang2022training} and high-quality conversational-style datasets using reinforcement learning from human feedback (RLHF) \cite{christiano2017deep, li2019human}.

Besides its dialogue ability, ChatGPT is being extensively studied by the researchers. Some literature focus on the ethical or other risks brought by the powerful LLMs like ChatGPT \cite{haque2022think, zhuo2023exploring, krugel2023moral}. chatGPT has also revolutionized the traditional fields like education \cite{basic2023better, kortemeyer2023could} and medicine \cite{susnjak2023applying,tu2023causal,nov2023putting}, and what will these industries advance with ChatGPT is analyzed. There are also research work investigating how ChatGPT performs in various natural language processing tasks. There are work testing ChatGPT on a wide range of NLP tasks, but usually a random test subset of the original benchmark tasks is tested \cite{qin2023chatgpt,bang2023multitask}. Other works focus on a specific type of tasks. For instance, \cite{ortega2023linguistic,susnjak2023applying} work on sentiment analysis, \cite{bian2023chatgpt} works on commonsense reasoning. The ChatGPT's ability of machine translation is examined in \cite{2023arXiv230108745J}. \cite{frieder2023mathematical} evaluates the mathematical capabilities of ChatGPT. Additionally, \cite{mitrovic2023chatgpt, guo2023close} investigate the differences between human-written and ChatGPT-generated contents in tasks like summarization and question answering, and the linguistic and grammatical features of ChatGPT's responses are analyzed.

\subsection{Information Extraction}

Information Extraction (IE) is a research topic of long history that aims to extract structured knowledge or factual information from unstructured texts \cite{yang2022survey}. The field of IE includes a wide range of tasks, such as named entity recognition \cite{das-etal-2022-container,landolsi2023information}, relation extraction (RE) \cite{2020arXiv200910680Z,li2022sequence}, event extraction \cite{hsu2022degree}, aspect-level sentiment analysis \cite{cheng57aspect}. Since the raise of pre-trained models like BERT \cite{devlin2018bert}, the performances on IE tasks have advanced greatly \cite{zhu-2021-mvp}. But one has to have different model structures for different fine-grained IE tasks, for instance, the SOTA nested NER models \cite{Zhang2022DeBiasFG} are different from those of discontinuous NER tasks \cite{zhang-etal-2022-de}. Recently, there is a trend that all the IE task should be solved by a unified paradigm, that is, Seq2Seq generation. \cite{2021arXiv210601223Y} proposes the framework of BartNER which solves all types of NER tasks with a BART model \cite{Lewis2019BARTDS}. UIE \cite{lu-etal-2022-unified} takes a step ahead and proposes to use prompts and a unified structural language to deal with many types of IE tasks with a single model checkpoint. 

With the rise of LLMs, the research field of IE is also under revolution, and this work contributes to the literature by conducting a deep investigation into the ChatGPT's capabilities in MedIE tasks. 

% NAT-NER: introduce non-autoregressive generation model into solving different types of NER tasks. 

\subsection{Medical natural language processing}

The developments in neural networks and natural language processing has advanced the field of medical natural language processing (MedNLP) \cite{2021arXiv211015803Z,hahn2020medical,zhu-etal-2021-discovering}. In the pre-BERT era, firstly, RNNs like LSTM/GRU are used for processing sequential medical data such as text and speech \cite{beeksma2019predicting}. Convolutional networks are also used for medical text classificaiton \cite{hughes2017medical}. The techniques of Graph neural networks are also explored for diagnose recommendations \cite{li2020graph}. In this period, many different model architectures are specially designed for better performances on a specific MedNLP task \cite{zhu-etal-2021-discovering,zhu2021autotrans,zhang2021automatic}. Since BERT \cite{devlin2018bert}, the pretrained language models (PLMs) become the deafult solution for MedNLP. In this stage, researcher becomes less interested in modifying the model architecture, but instead trying to pretrain or further pretrain a PLM from the open domain to the medical domain \cite{guo2021global,zhu-2021-mvp,pubmedbert}.

With the wide study of LLMs, the field of MedNLP is also being revolutionized. There are already works on adapting LLM backbones to the medical domain question answering \cite{zhu2023ChatMed}. And \cite{zhu2023PromptCBLUE} propose PromptCBLUE, a prompt learning based benchmark dataset for examing the LLMs' ability in MedNLP tasks. This work investigates MedIE capabilities in ChatGPT, and provide a deeper understanding of LLMs for future MedNLP research.

\section{ChatGPT for MedIE}

In this section, we first briefly introduce six fine-grained MedIE tasks, then we present how to evaluate ChatGPT's performances on these tasks with five different metric dimensions. The evaluation procedure involves annotations by the domain experts. 

\subsection{MedIE tasks}

MedIE involves a wide range of tasks which need to extract structured factual information from unstructured medical texts, such as entity, and relation, and event. In this research, we conduct our analysis on the following 4 MedIE tasks: 

\textbf{Named entity recognition} (NER) \quad NER aims to first identify the candidate medical entities, and then classify their types. Medical NER is known for its complexity, since it usually involves nested entities or discontinous entities \cite{2021arXiv210601223Y}.  

\textbf{Triple extraction} (TE) \quad This task identifies the target entities and the relations between each pair of entities jointly. The medical knowledge is expressed not by entities, but their relations. For example, a medicine can cure a diease, or a symptom is caused by a disease. This task is generally more complex than merely identifying entities or classifying the relations of a given pair of entities.  

\textbf{Clinical Event Extraction} (CEE) \quad This task is an example of event detection tasks \cite{2022arXiv221003419D} in the medical field. Same as the general event detection tasks, medical CEE task requires a model to identify the event triggers, classify the event types and extracts different arguments and categorize their roles with respect to the target events. 

% A branch of literature address this task as a classification task. But due to limited dataset size and the extreme long-tail property, this approach can not achieve good enough performances. 
\textbf{ICD encoding} \quad This task aims to map the diagnosis terms (query terms) written by doctors to standardized disease terms (target terms) according to a certain standard \cite{huang-etal-2022-plm}. The number of standardized terms may exceed 10 thouand. Recently, this task is modeled by a system where a small set of candidate terms are firstly retrieved and then a ranking model scores and ranks the relevances of each query-target term pair \cite{2021arXiv210710652K}. The standard system adopted in this task is usually the International Statistical Classification of Diseases and Related Health Problems 10th Revision (ICD-10)\footnote{https://www.who.int/standards/classifications/classification-of-diseases}. ICD-10 has more than 30 thousand disease terms, thus it is prohibitive to feed all the disease terms into ChatGPT. We first retrieve candidate target terms for each query term using BM25 \cite{robertson2009probabilistic}, then ask ChatGPT to be a ranking model and choose the final target terms among the candidates. ICD-10 has different versions in different countries, and in this work we adopt the ICD-10 Beijing Clinical Trial (version v601)\footnote{http://www.cips-chip.org.cn/2021/eval3}. We will refer to this Chinese version of ICD-10 as ICD-10-Beijing. 

% The standard system adopted in this task is usually the International Statistical Classification of Diseases and Related Health Problems 10th Revision (ICD-10) \footnote{xxxx} . ICD-10 is an international general disease classification method developed by the WHO. It classifies diseases into categories according to the specificity of the disease's etiology, clinical manifestations, and analytical location. ICD-10 has different versions in different countries, and in this work we adopt the ICD-10 Beijing Clinical Trial (version v601)\footnote{xxx, http://www.cips-chip.org.cn/2021/eval3}. We will refer to this Chinese version of ICD-10 as ICD-10-Beijing. 

% ICD-10-Beijing has more than 40 thousand disease terms, thus it is prohibitive to feed all the disease terms into ChatGPT. We first retrieve candidate target terms for each query term using BM25 (xxx, ), then ask ChatGPT to be a ranking model and choose the final target terms among the candidates. 

In summary, every task needs LLMs’ unique ability to perform well. It is worth to explore the performances on these fine-grained MedIE tasks on ChatGPT.

\subsection{Composition of task prompts}

To comprehensively evaluate the overall perfor-mance of ChatGPT on IE tasks, we ask ChatGPT to generate the responses under the closed MedIE setting. In the closed MedIE setting \cite{niklaus2018survey}, the labels of the entities/relations/events are pre-defined and a model to extract information accordingly and no other information should be extracted. The closed MedIE paradigm is the most commomly used in previous works, which uses the task-specific dataset with supervised learn-ing paradigm to finetune a model like BERT \cite{devlin2018bert}. 

For ChatGPT, as we can not directly fine-tune the model parameters, we evaluate the ChatGPT’s ability to conduct MedIE with task prompts. Specifically, the task prompt consists of the following parts: 
\begin{itemize}
	\item Task descriptions, describing the objective of the task, and the steps that should be taken to achieve the objective.
	\item label sets, containing all the candidate labels in this task. This part also includes the explanations of the labels.    
	\item Output formats, regulating the formats of the responses. Since MedIE tasks extract structured information from the un-structured texts, what fields are in the output results should be specified. And we need to specify the output formats so that we can evaluate the performance of ChatGPT. 
	\item Demonstrations: with a few input-output examples of an unseen task, the LLMs can learn from the examples and perform the task without optimizing any parameters \cite{li2023unified}.
	\item The input text of the sample that we want the model to make prediction on.
\end{itemize}

The prompts is in the same language as the MedIE task. Due to limited length, we only present the prompt for CMeEE-v2 \cite{hongying2021building}, a Chinese NER task, in Table \ref{tab:example}. The left size is the prompt contents in Chinese, and the right size is their English translation.

\begin{CJK*}{UTF8}{gbsn}

\begin{table*}[]
	\caption{The task prompt for the medical NER task CMeEE-v2 \cite{hongying2021building}. The left is the prompt in Chinese, and the right size is its English translation }
	\label{tab:example}
	\centering
	\resizebox{0.99\textwidth}{!}{\begin{tabular}{ll}
		\toprule
		\textit{任务名称：}  & \textit{Task name} \\ 
		\midrule
		医疗文本实体抽取      &  NER for medical texts \\ \hline
		\textit{任务描述：}    &   \textit{Task description: } \\ \hline
		\begin{tabular}[l]{@{}l@{}}对于给定的一组纯医学文本文档，任务的目标是识别并抽取出与医学临床相关的实体，\\ \quad 并将他们归类到预先定义好的类别。 \end{tabular}  &   \\ \midrule
		\textit{实体类型含义解释：}   &  \textit{Explanations of the entity labels: } \\ \hline
		\begin{tabular}[l]{@{}l@{}}1. 疾病：指导致病人处于非健康状态的原因或者医生对病人做出的诊断，并且是能够被治疗的。\\ \quad 包括疾病或综合征、中毒或受伤、器官或细胞受损等\\ 2. 临床表现：疾病的表现，泛指患者不适感觉以及通过检查得知的异常表现。主要包括症状、体征\\ 3. 医疗程序：泛指为诊断或治疗所采取的措施、方法及过程。主要包括检查程序、治疗或预防程序等\\ 4. 医疗设备：泛指为诊断或治疗所使用的工具、器具、仪器等。主要包括检查设备、治疗设备\\ 5. 药物：指用来预防、治疗及诊断疾病的物质，包括临床药物、抗生素等\\ 6. 医学检验项目：指检查涉及到的体液检查项目、重要生理指标以及其他检查项目，主要针对人体而言，\\ \quad 是能够通过设备或实验检测出的项目，并且是能够被量化，有其对应的测量值或指标值\\ 7. 身体解剖部位：泛指细胞、组织、及位于人体特定区域的由细小物质成分组合而成的结构、\\ \quad 器官、系统、肢体，另外包括身体产生或解剖身体产生的物质等\\ \quad 8. 科室：指医院或医疗机构所设有的科室\\ 9. 微生物类：包括细菌、病毒、真菌以及一些小型的原生生物、\\ \quad 显微藻类等在内的一大类生物群体，另外包括微生物类产生的毒素、激素、酶等\end{tabular}     &  \begin{tabular}[l]{@{}l@{}} 1. Disease: refers to the cause of the patient's unhealthy state or the doctor's diagnosis of the patient, \\ \quad and it can be treated. Including diseases or syndromes, poisoning or injury, organ or cell damage, etc. \\ 2. Clinical manifestations: manifestations of diseases, generally referring to the patient's discomfort \\ \quad and abnormal manifestations learned through examination. It mainly includes symptoms and signs \\ 3. Medical procedure: generally refers to the measures, methods and processes adopted for diagnosis or \\ \quad treatment. It mainly includes inspection procedures, treatment or prevention procedures, etc. \\ 4. Medical equipment: generally refers to tools, appliances, instruments, etc. used for diagnosis or treatment.\\  \quad Mainly including inspection equipment, treatment equipment \\5. Drugs: refer to substances used to prevent, treat and diagnose diseases, \\  \quad including clinical drugs, antibiotics, etc.\\ 6. Medical inspection items: refer to body fluid inspection items, important physiological indicators \\ \quad and other inspection items involved in the inspection, mainly for the human body, \\  \quad and is an item that can be detected by equipment or experiments, and can be quantified, \\  \quad and has its corresponding measurement value or index value\\ 7. Body anatomical parts: generally refer to cells, tissues, and specific areas located in the human \\ \quad body structures, organs, systems, and limbs composed of small material components, including \\ \quad substances produced by the body or dissected by the body, etc. \\ 8. Department: refers to the department set up by a hospital or medical institution\\ 9. Microorganisms: a large group of organisms including bacteria, viruses, fungi, some small protists, \\ \quad microscopic algae, etc., and toxins, hormones, enzymes, etc. produced by microorganisms \end{tabular} \\ 
		\midrule
		\textit{示例：}    &  -  \\ \hline
		\begin{tabular}[l]{@{}l@{}}示例1\\ input: 对儿童SARST细胞亚群的研究表明，与成人SARS相比，儿童细胞下降不明显，\\ \quad 证明上述推测成立。\\ target: {[}\{'实体类型': '身体解剖部位', '实体名称': 'SARST细胞亚群'\}, \\ \quad \{'实体类型': '疾病', '实体名称': '成人SARS'\}, \\ \quad \{'实体类型': '临床表现', '实体名称': '细胞下降'\}, \\ \quad \{'实体类型': '身体解剖部位', '实体名称': '细胞'\}{]}\end{tabular} &  \begin{tabular}[l]{@{}l@{}}  Example 1\\ input: The research on children's SARST cell subsets shows that compared with adult SARS,\\ \quad the children's cells  are not significantly reduced, proves that the above speculation is true. \\ target: {[}\{'Entity Type': 'Body Anatomy', 'Entity Name': 'SARST Cell Subgroup'\}, \\ \quad \{'Entity Type': 'Disease', ' Entity name': 'Adult SARS'\}, \\ \quad \{'Entity type': 'Clinical manifestation', 'Entity name': 'Cell decline'\}, \\ \quad \{'Entity type': 'Body Anatomy', 'Entity Name': 'Cell'\}{]}  \end{tabular} \\ 
		\midrule
		\textit{输出格式规定：}    &   \textit{Output format：}  \\ 
		\midrule
		\begin{tabular}[l]{@{}l@{}}1. 本任务的输出一个字典的列表，必须符合json格式。\\ 2. 每个字典代表一个医学实体，包含\textbackslash{}"实体类型\textbackslash{}", \textbackslash{}"实体名称\textbackslash{}"字段，每个字段的取值都为字符串。\\ 3. 实体名称：即为医学实体在给定的文本中的出现形式，实体名称可以包含必要的标点符号，\\ \quad 即可以是一个词、短语或句子。“临床表现”实体类别中允许嵌套，即该类型实体的名称内部可能\\ \quad 包含其他八类实体。除了“临床表现”实体之外的医学实体，在抽取其实体名称时遵循“最大单位 \\ \quad 标注法”，即如果一个实体名称里包含其他的实体，只需要将最长的实体标注出来\\ 4. 注意：你需要一步步的思考，然后形成医学实体列表。\end{tabular}       &  	\begin{tabular}[l]{@{}l@{}}   1. The output of this task is a list of dictionaries, which must conform to the json format. \\ 2. Each dictionary represents a medical entity, including \textbackslash{}"entity type\textbackslash{}", \textbackslash{}"entity name\textbackslash{}" fields, \\ \quad and the value of each field is a string . \\ 3. Entity name: It is the appearance form of the medical entity in the given text, the entity name can \\ \quad contain necessary punctuation marks,  and it can be a word, phrase or sentence. Nesting is allowed in \\ \quad the "clinical manifestation" entity category, that is, the name of this type of entity may contain other eight \\ \quad types of entities. Medical entities other than "clinical manifestation" entities follow the "maximum unit \\ \quad labeling method" when extracting their entity names, that is, if an entity name contains other entities, \\ \quad only the longest entity needs to be marked out \\ 4. Note: You need to think step by step, and then form a list of medical entities.
		  \end{tabular}   \\ 
		\midrule
		\textit{输入:}     &  \textit{Current input text:} \\ 
		\midrule
	\begin{tabular}[l]{@{}l@{}} input:【发病率】自1952年Bruton报告首例先天性无丙种球蛋白血症（congenitalagammaglobulinemia）\\以来，全球报道的PID病例已愈万例，但其总发病率尚无确切资料。\end{tabular}    &  \begin{tabular}[l]{@{}l@{}}  input: [Incidence] Since Bruton reported the first case of congenital agammaglobulinemia \\ \quad (congenital agammaglobulinemia) in 1952, more than 10,000 cases of PID have been reported \\ \quad worldwide, but there is no definite information on the total incidence.  \end{tabular}  \\ 
		\bottomrule
	\end{tabular}
}
\end{table*}

\end{CJK*}

\subsection{Evaluation protocols}

% By examining its performance from different aspects, we seek to provide a detailed understanding of ChatGPT’s capability on the downstream MedIE tasks.
% The explainability of deep learning models is crucial for the application in real-world scenarios (xxx, ).

To systemically and comprehensively evaluate the ChatGPT's capabilities on the MedIE tasks, we evalute ChatGPT's responses under the five dimensions:
\begin{itemize}
	\item \textbf{Performance}. \quad One import objective of this work is to comprehensively evaluate the overall performance of ChatGPT against the ground truth on various MedIE tasks and compare it with other popular fine-tuned models. 
	
	\item \textbf{Explainability} \quad  In this work, we examine the explainability of ChatGPT by asking it to provide detailed and accurate explainations of its reasoning processes for conducting information extraction on a given task. We collect the explainations from ChatGPT at two levels: (a) sample level (denoted as \emph{sample\_level\_explain}), that is, we ask ChatGPT to provide an overall explaination of the whole input's extraction results. (b) instance level, (denoted as \emph{instance\_level\_explain}), that is, explanation is provided for each extracted instance (for example, each entity mention). 
	
	\item \textbf{Faithfulness} \quad The faithfulness of LLMs is important to ensure their trustworthiness \cite{2023arXiv230410513Z}. For evaluating ChatGPT's faithfulness, we examine two aspects: (a) whether the responses made by ChatGPT follows the task instructions and extract information from the input. This aspect of faithfulness is denoted as \emph{Instruct\_following} (b) whether the explainations made by the ChatGPT are faithful to the original sample input. This aspect is denoted as \emph{faithful\_reasoning}.
	\item \textbf{Confidence} \quad We aim to examine whether ChatGPT has over-confidence phenomenon. 
	\item \textbf{Uncertainty} \quad Since ChatGPT generates responses using the top-p sampling \cite{2019arXiv190409751H} decoding algorithm, the prediction uncertainty of ChatGPT is an important aspect that should be considered when applying ChatGPT to different applications. In this work, we measure the prediction uncertainty of ChatGPT by querying the same sample to ChatGPT API for 5 times and measure the differences among different runs.      
\end{itemize}
% With the advances of neural networks, especially LLMs, the industry and users will become increasingly more reliant on the artificial intellegence models. However, there are valid concerns that we are applying black-box algorithms to provide solutions to problems that could result in catastrophic failure. 

\section{Performances}

\subsection{Experimental Setups}

To ensure a comprehensive evaluation of ChatGPT's abilities on the MedIE task, we conduct experiments on a diverse range of MedIE tasks, including 4 different tasks spanning 6 datasets: (a) for medical NER tasks, we include three widely studied benchmark datasets, ShARe13 \cite{Mowery2013Task1S}, CADEC \cite{Karimi2015CadecAC}, CMeEE-V2 \cite{hongying2021building}. (b) for triple extraction task, we include the CMeIE-v2 dataset \cite{2020CMeIE}; (c) The CDEE dataset \cite{zhang-etal-2022-cblue} is a benchmark Chinese clinical event extraction task. (d) for the ICD-coding task, the CHIP-CDN task \cite{zhang-etal-2022-cblue} is adopted for evaluation. 

% . The ShaRE-2013 and CMeEE-V2 dataset contains nested entities, while the CADEC dataset contains discontinuous entities. The above three datasets represents the complex NER bechmarks and are widely studied in the literature for general NER approaches. (b) for triple extraction task, we include the CMeIE-v2 dataset (xxx). 

We use the official OpenAI API (with the \emph{gpt-3.5-turbo} engine) to obtain the ChatGPT responses. To prevent any biases caused by historical chats, we cleared the conversation after generating each response. For the ShaRE-2013 and CADEC datasets, we ask ChatGPT to evaluate on the test set. For the rest tasks, the test set ground truths are not publicly available, so we conduct evaluations on the development set. Otherwise specified, we choose the first two samples from the training set as demonstrations. 

For comparison, we compare ChatGPT with a series of popular baselines: (1) BERT fine-tuning. For all the six task, we adopt the bi-affine attention module as the prediction module, following \cite{dozat2016deep} (2) UIE \cite{lu-etal-2022-unified}. (3) The state-of-the-art (SOTA) methods on each dataset, as reported in Table \ref{tab:main}.

For each task, we report the instance level strict F1-score. Here, an instance is a piece of structured information/knowledge from the un-structured text. For example, the entity mention and entity type constitutes an instance for the NER tasks. For the ICD coding task, since a diagnosis term from the doctor may correspond to more than one ICD standard terms, an instance in this task is defined as a ICD-10 standard disease term. The F1-score metric we adopt is strict, in the sense that the true positive number will add one only when all the keys in an ground-truth instance is correctly predicted.

Due to the time-consuming nature of obtaining responses from domain experts, we randomly select 200 samples for each task for manual annotations. For each sample, we ask the human experts to annotate \emph{sample\_level\_explain}, \emph{instance\_level\_explain} and \emph{faithfull\_reasoning} as boolean values. 

% (a) whether the \emph{sample\_level\_explain} is reasonable; (b) whether the \emph{instance\_level\_explain} is reasonable; (c) \emph{faithfull\_reasoning}, that is, the extracted instances and explainations are faithful to the original input and not hullucinated by the LLM. 

% (c) \emph{instruct\_following}, that is, whether ChatGPT follows instructions and output the results in the format we specified; 

\subsection{Performance of ChatGPT}

% \multirow{2}*{Source}
% \multicolumn{4}{c}{Target}
\begin{table*}
\caption{The performances of ChatGPT and several baseline models on the six MedIE datasets. We report the average performance on the whole dev set over 5 different runs (with standard deviations reported in brackets). All the baseline models are re-implemented using official open-source codes. }
\label{tab:main} 
\centering
\resizebox{0.85\textwidth}{!}{
\begin{tabular}{cccccc}
\hline
\textbf{MedIE Tasks} & \textbf{Dataset}   &  \textbf{BERT}  &  \textbf{UIE}  & \textbf{SOTA}   &  \textbf{ChatGPT}      \\
\hline
\multirow{3}*{Named Entity Recognition}  &  ShARe13  &  76.5 ($\pm0.2$)  &  78.8 ($\pm0.4$) &  79.6 ($\pm0.7$) \cite{zhang-etal-2022-de}  &   23.5 ($\pm1.2$)  \\
&  CADEC     &    70.3 ($\pm0.4$) &  70.6 ($\pm0.6$) &  71.6 ($\pm0.4$) \cite{zhang-etal-2022-de}   &   19.7 ($\pm1.5$) \\
&  CMeEE-v2  &   66.2 ($\pm0.3$)  &  66.7 ($\pm0.4$) &   68.5 ($\pm0.3$) \cite{du-etal-2022-mrc}   &  39.8 ($\pm2.1$)  \\
\hline
\multirow{1}*{Triple Extraction}  &  CMeIE-v2   &  58.6 ($\pm0.1$) &  57.4 ($\pm0.3$)  &   60.1 ($\pm0.5$) \cite{2021arXiv211007244W} &  9.9 ($\pm0.8$)  \\
\hline
\multirow{1}*{Event Extraction}  &  CHIP-CDEE  &  62.4 ($\pm0.5$) &  64.1 ($\pm0.5$)  &  65.8 ($\pm0.4$) \cite{hsu-etal-2022-degree}  &  26.1 ($\pm1.8$)  \\
\hline
\multirow{1}*{ICD Coding}  &  CHIP-CDN  &   78.9 ($\pm0.3$)  &  82.3 ($\pm0.2$)   &   84.9 ($\pm0.3$) \cite{zhu2023PromptCBLUE}  &   30.9 ($\pm1.1$) \\
\hline
\end{tabular}}
\end{table*}

In this subsection, we report the performances of different models including ChatGPT on the six benchmark MedIE tasks in Table \ref{tab:main}. It is clear from Table \ref{tab:main} that ChatGPT’s performance is not comparable to that of the fully fine-tuned baseline models or the SOTA methods in any of the MedIE tasks. This is not surprising given that ChatGPT does not fine-tune its parameters to adapt to the task, and asking it for the prediction with only prompts is an extreme few-shot scenario. In comparison, the other compared models are fully fine-tuned on task-specific datasets under a supervised learning paradigm. Another reason may be that the meaning of some labels may not be easy to understand even though we have provide explanations, thereby negatively impact the performance. In addition, we can see that due to the top-p sampling strategy, the standard deviation (i.e., uncertainty) in performance scores for ChatGPT is generally higher than the baseline models.

The results in Table \ref{tab:main} indicate that ChatGPT performs slightly better on relatively simple MedIE tasks but have difficulty in dealing with complex and challenging tasks. For example, the ICD coding task only involves matching the diagnoses term from medical records to the candidate standard terms from the ICD-10 vocabulary, and no further contextual understanding are required. And since in the CMeEE task, even though there are nested entities, more than 90\% of the entities are flat. Thus, ChatGPT can achieve F1-scores of around 30 in these two tasks. However, in the more complex tasks like CMeIE-v2, ChatGPT struggles as it needs to firstly identify the entities that exist in the input and then identify which entity pairs have meaningful relationships. In summary, ChatGPT’s performance varies based on the complexity of the task.

Note that in our results, the ChatGPT falls behind the fine-tuned baseline models. This conclusion is inconsistent with the previous studies \cite{2023arXiv230206476Q,gao2023exploring,2023arXiv230411633L}, which show that ChatGPT can achieve desirable performance in a wide range of NLP tasks including IE tasks. One possible explanation for the difference in conclusions is that we report the performance of the entire dev or test set for each task in our study, while prior studies reported on a very small set of test samples drawn at random, which may have substantial variance. In addition, the MedIE tasks requires domain knowledges and are more complex and challenging than the tasks in the general domain \cite{pubmedbert}.

\subsection{Other evaluation dimensions}

While it is important to obtain ChatGPT's performance in evaluation, it is equally important to evaluate its capabilities from a wide range of dimensions that could offer important insights for future research. In this subsection, we analyze the explainability, faithfulness and confidence to give better insights into the ChatGPT’s abilities. 

\subsubsection{Explainability} 

% Explainability is an essential factor for the modern neural networks like LLMs, as it allows users to understand how the model arrives at its predictions and be comfortable using the applications supported by LLMs. In this work, we investigate whether ChatGPT could provide a reasonable explanation for its outputs. To be specific, we ask ChatGPT to provide reasons for its prediction at both the sample level (\emph{sample\_level\_explain}) and instance level (\emph{instance\_level\_explain}). The generated reasons are then evaluated for their appropriateness and reasonableness by three domain experts. The resulting evaluations are in boolean values, where True means the reasons provided by ChatGPT is valid while False means not. If two or more of the expert vote True, then the explaination provided by the ChatGPT is considered reasonable. For \emph{instance\_level\_explain}, we consider both the correctly predicted instances and the wrongfully predicted instances to ensure a valid evaluation of ChatGPT's explainability ability. We randomly select around 300 samples or 600 input-instance pairs from each dataset for human annotation.  

Explainability is an essential factor for the modern neural networks like LLMs, as it allows users to understand how the model arrives at its predictions and be comfortable using the applications supported by LLMs. To evaluate \emph{instance\_level\_explain}, we only consider the correctly predicted instances to ensure a valid evaluation of ChatGPT's explainability ability. The ratio of samples/instances with reasonable explanations (denoted as R-score) for each task is reported in Table \ref{tab:explanations}, and the following take-aways can be drawn. Firstly, human annotators approve the reasons given by ChatGPT, with the majority of datasets achieving a R-score of over 750\%. The above results demonstrate that ChatGPT gives high-quality explanation for its prediction. Secondly, we find that ChatGPT seems to be over-confident in its explanations, since on the wrong predictions ChatGPT can also provide explanations with confident tones and it can not reflect on its mistakes.

\begin{table}
\caption{The R-scores for the explanations provded by the ChatGPT of its prediction both at the sample level and instance level. }
\label{tab:explanations} 
\centering
\resizebox{0.48\textwidth}{!}{
\begin{tabular}{ccc}
\hline
Dataset &  sample-level R-score   &  instance-level R-score  \\
\hline
ShARe13  &  76\%  &  91\%  \\
CADEC  &  77\%  &  89\%  \\
CMeEE-v2  &  81\%  &  93\%   \\
CMeIE-v2  &  79\%  &  95\% \\
CHIP-CDEE  &   76\%  &  92\%   \\
CHIP-CDN  &   77\%  &  88\%   \\
\hline
\end{tabular}}
\end{table}

\subsubsection{Confidence} 

In this part, we investigate the level of confidence provided by ChatGPT for both the correctly and incorrectly predicted instances. Our objective is to investigate whether ChatGPT can provide reasonable prediction confidence scores for its predictions, thus reducing the risk of over-confidence or mis-interpretation. In Table \ref{tab:confidence}, we present the average confidence scores of correct and incorrect predictions from ChatGPT, referred to CC-score and IC-score, respectively. The results in Table \ref{tab:confidence} reveal that ChatGPT has high confidence levels in their correct and incorrect predictions. This is in line with the previous literature reporting that large scale neural network models are over-confident \cite{wang2021rethinking}. This overconfidence may present risk of misguidances to users. 

In addition, we find two additional observations: (a) there are no significant gaps between the CC-score and IC-score, indicating the need for careful evaluation of ChatGPT's outputs since high confidence scores does not mean the predictions are more likely to be correct. (b) The confidence scores have lower standard deviations. In fact, many of the samples have the same confidence scores. The above two observations reveal the un-reliability of ChatGPT's confidence scores.

% (both CC-score which is the average confidence score for the correctly predicted instances, and IC-score which is the average confidence score for incorrectly predicted instances)
\begin{table}
\caption{The average confidence scores estimated by ChatGPT for its predicted instance. The standard devaitions of confidence scores are reported in bracket. }
\label{tab:confidence} 
\centering
\resizebox{0.36\textwidth}{!}{
\begin{tabular}{ccc}
\hline
Dataset &  CC-score  &   IC-score   \\
\hline
ShARe13  &  81.4 ($\pm$1.2)   &   81.1 ($\pm$0.9)  \\
CADEC  &   79.8 ($\pm$1.0)   &   80.1 ($\pm$1.7)  \\
CMeEE-v2  &   82.2 ($\pm$0.8)   &   81.9 ($\pm$1.1)   \\
CMeIE-v2  &    79.6 ($\pm$1.3)   &   78.9 ($\pm$1.1)   \\
CHIP-CDEE  &    79.3 ($\pm$1.5)   &   79.1 ($\pm$1.3)   \\
CHIP-CDN  &    81.5 ($\pm$0.9)   &   80.7 ($\pm$0.8)   \\
\hline
\end{tabular}}
\end{table}

\subsubsection{Faithfulness} 

Assessing the faithfulness of ChatGPT is critical in developing a trustworthy information extraction model. Recent work argue that LLMs could provide hullucinated information to users, which could potentially misguide the users' decision making \cite{zhang2023benchmarking}. In this work, we measure the faithfulness of ChatGPT on the MedIE tasks in the following two aspects: (a) \emph{Instruct\_following}, that is, whether the responses made by ChatGPT follows the task instructions and extract information strictly following the inputs and the given label sets or candidate sets. (b) \emph{faithful\_reasoning}, that is, whether the explanations given by ChatGPT are faithful to the original sample inputs.  

Table \ref{tab:faithful} reports the ratio of the annotated samples in which the \emph{Instruct\_following} or \emph{faithful\_reasoning} key is true. We can see that on most of the MedIE tasks, the \emph{Instruct\_following} and \emph{faithful\_reasoning} scores exceed 80\%, showing that ChatGPT can follow the task instructions and its explanations are regarded as reliable according to the inputs. This result also demonstrate that the performance gap from ChatGPT to the fine-tuned baselines are not caused by generating labels that are not in the label sets.

\begin{table}
	\caption{The ratio of the annotated samples in which the \emph{Instruct\_following} or \emph{faithful\_reasoning} key is true}
	\label{tab:faithful} 
	\centering
	\resizebox{0.48\textwidth}{!}{
		\begin{tabular}{ccc}
			\hline
			Dataset &  \emph{Instruct\_following}  &   \emph{faithful\_reasoning}   \\
			\hline
			ShARe13  &  86\%   &  96\%  \\
			CADEC  &  87\%  &  93\%   \\
			CMeEE-v2  &  89\%  &  94\%  \\
			CMeIE-v2  &   83\%  &   91\% \\
			CHIP-CDEE  &   89\%   &  95\%  \\
			CHIP-CDN  &   82\%  &   93\% \\
			\hline
	\end{tabular}}
\end{table}

\section{Conclusions}

In this paper, we systematically analyze the ChatGPT's performance, uncertainty, explainability, over-confidence and faithfulness on medical information extraction (MedIE) tasks. With carefully designed task prompts, we collect ChatGPT's responses via OpenAI APIs. Expert annotations are included for reliable evaluation. On the 6 MedIE tasks, we find that: (a) ChatGPT's performance is not satisfying compared to fully fine-tuned baseline models. (b) ChatGPT can provide valid explanations for its predictions. (c) The over-confidence issue is observed in ChatGPT, indicating low calibration and risk for application. (d) ChatGPT exhibits faithfulness to the input texts and task prompts. (e) The uncertainty caused by the randomness in response decoding results in high uncertainty in the predicted results. We hope that our work could be the basis for future research on applying LLMs like ChatGPT for information extraction.

\bibliographystyle{IEEEbib}
\bibliography{refs}

\end{document}